\documentclass[10pt,twocolumn,letterpaper]{article}

\usepackage[pagenumbers]{cvpr} 
\usepackage[table]{xcolor}
\newcommand{\xmarkg}{\textcolor{lightgray}{\ding{55}}\xspace}%
\definecolor{myrowgray}{RGB}{240,240,240}

\usepackage{multirow}
\usepackage{caption}
\usepackage{makecell}
\usepackage{comment}
\usepackage{float}
\usepackage{cuted}
\usepackage{caption}
\usepackage{pifont}
\newcommand{\xmark}{\textcolor{lightgray}{\ding{55}}}
\newcommand{\cmark}{{\ding{51}}}

\usepackage[misc]{ifsym}
\definecolor{cvprblue}{rgb}{0.21,0.49,0.74}
\usepackage[pagebackref,breaklinks,colorlinks,allcolors=cvprblue]{hyperref}
\usepackage[table]{xcolor}
\usepackage{float}
\def\paperID{8954} 
\def\confName{CVPR}
\def\confYear{2026}
\title{ROSE: Retrieval-Oriented Segmentation Enhancement}

\author{
Song Tang\footnotemark[1]
\quad
Guangquan Jie\footnotemark[1]
\quad
Henghui Ding~\!$^{\textrm{\Letter}}$
\quad
Yu-Gang Jiang
\vspace{.6mm}
\\
Fudan University, China
\vspace{.6mm}
\\
\href{https://henghuiding.com/ROSE/}{https://henghuiding.com/ROSE/}
}

\begin{document}

\twocolumn[{%
\renewcommand\twocolumn[1][]{#1}%
\maketitle
\vspace{-3.6mm}
\begin{center}
\centering
\captionsetup{type=figure}
\includegraphics[width=1\linewidth]{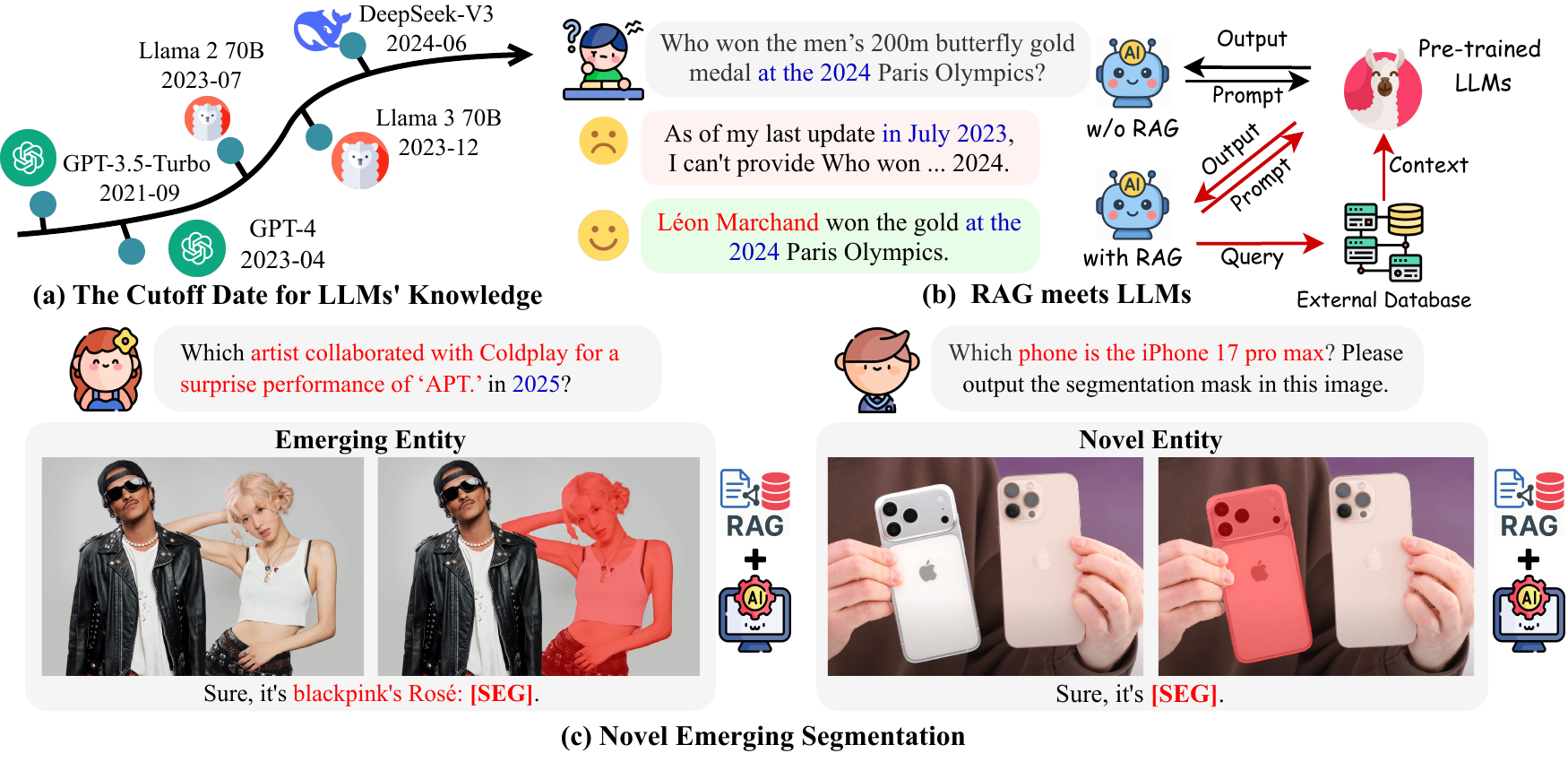}
\vspace{-6.6mm}
    \captionof{figure}{\textbf{Novel Emerging Segmentation.} (a) The cutoff date for Large Language Models'~(LLMs) training data limits their knowledge of recent events. (b) Retrieval-Augmented Generation~(RAG) enhances LLMs by retrieving up-to-date information from external databases.
    (c) Our task focuses on segmenting novel entities unrecognizable by existing models due to their absence in training data, and emerging entities that exist in the models' knowledge but require up-to-date external information.}
\vspace{6.8mm}
\label{fig:task}
\end{center}
}]

\renewcommand{\thefootnote}{\fnsymbol{footnote}}
\footnotetext[1]{Equal contribution.}
\footnotetext[0]{${\textrm{\Letter}}$ Henghui Ding (henghui.ding@gmail.com) is the corresponding author with the Institute of Big Data, College of Computer Science and Artificial Intelligence, Fudan University, Shanghai, China.}

\begin{abstract}
Existing segmentation models based on multimodal large language models (MLLMs), such as LISA, often struggle with novel or emerging entities due to their inability to incorporate up-to-date knowledge.
To address this challenge, we introduce the Novel Emerging Segmentation Task (NEST), which focuses on segmenting (i) novel entities that MLLMs fail to recognize due to their absence from training data, and (ii) emerging entities that exist within the model's knowledge but demand up-to-date external information for accurate recognition.
To support the study of NEST, we construct a NEST benchmark using an automated pipeline that generates news-related data samples for comprehensive evaluation.
Additionally, we propose \textbf{ROSE}: \textbf{R}etrieval-\textbf{O}riented \textbf{S}egmentation \textbf{E}nhancement, a plug-and-play framework designed to augment any MLLM-based segmentation model.
ROSE comprises four key components.
First, an Internet Retrieval-Augmented Generation module is introduced to employ user-provided multimodal inputs to retrieve real-time web information.
Then, a Textual Prompt Enhancer enriches the model with up-to-date information and rich background knowledge, improving the model's perception ability for emerging entities.
Furthermore, a Visual Prompt Enhancer is proposed to compensate for MLLMs' lack of exposure to novel entities by leveraging internet-sourced images.
To maintain efficiency, a WebSense module is introduced to intelligently decide when to invoke retrieval mechanisms based on user input.
Experimental results demonstrate that ROSE significantly boosts performance on the NEST benchmark, outperforming a strong Gemini-2.0 Flash-based retrieval baseline by 19.2\% in gIoU.
\end{abstract}

\vspace{-3.5mm}
\section{Introduction}
\label{sec:intro}

Segmentation models based on multimodal large language models~(MLLMs), \eg,~LISA~\cite{lisa}, SESAME~\cite{sesame}, and READ~\cite{read}, leverage MLLMs'~\cite{llava,internvl,gpt4v,gemini} reasoning abilities and world knowledge to address reasoning segmentation and achieve zero-shot capabilities.
For example, when given an instruction ``\textit{Please segment the founder of SpaceX}'', these models can identify the correct person in the image using MLLMs' knowledge.
However, MLLMs require substantial computational resources for data collection, cleaning, and training, making frequent updates impractical.
As a result, MLLM-based models struggle to incorporate newly emerging information.
Since knowledge in the real world evolves rapidly, this limitation leads to failure cases in segmentation tasks involving novel or recently emerged entities.
For example, LISA cannot accurately segment the current U.S. President due to its knowledge cutoff in 2023.
To address this, it is promising to augment MLLM-based segmentation methods with Retrieval-Augmented Generation~(RAG) techniques, as shown in Fig.~\ref{fig:task} (b), enabling the model to effectively access and leverage up-to-date knowledge during inference.

In this work, we introduce a new segmentation task, \textbf{Novel Emerging Segmentation Task (NEST)}, to evaluate models' ability on segmenting novel and emerging entities, as shown in Fig.~\ref{fig:task}~(c). NEST requires generating binary segmentation masks based on user queries, with a focus on segmenting novel and emerging entities that are either unseen during training or require up-to-date knowledge for accurate interpretation.
Novel entities refer to objects entirely absent from the MLLMs' training data.
For example, models like LLaMA 3~\cite{llama3}, which have a knowledge cutoff at the end of 2023, are unable to recognize products such as the iPhone 17 pro max, which was released in 2025.
Emerging entities, on the other hand, are included in the model’s prior knowledge but evolve over time and require current context for accurate segmentation. For example, while LISA can segment Joe Biden and Donald Trump individually, it may fail to identify the current U.S. President due to outdated knowledge.
Successfully addressing the NEST demands the following key capabilities:
1) retrieving up-to-date knowledge from the internet;
2) recognizing previously unseen entities;
3) applying the knowledge retrieved from the internet to generate accurate segmentation within visual inputs.

To support the study of novel emerging segmentation, we construct the \textbf{NEST benchmark}, containing over 1,500 image-question-answer-mask pairs. These samples are collected through an automated pipeline that continuously retrieves and updates the latest image-news pairs from the web.
This automated pipeline provides a scalable and practical way for generating diverse and timely evaluation data tailored to novel and emerging entities. It enables continuous assessment of model performance as new concepts and objects emerge in the open world.

To tackle novel emerging segmentation, we propose \textbf{R}etrieval-\textbf{O}riented \textbf{S}egmentation \textbf{E}nhancement (\textbf{ROSE}), a plug-and-play method that can be integrated with any MLLM-based segmentation model (\eg, LISA~\cite{lisa}).
ROSE consists of four key components. First, the Internet Retrieval-Augmented Generation (IRAG) module enhances segmentation by retrieving real-time information from the web based on user-provided multimodal inputs. Second, the Textual Prompt Enhancer (TPE) supplements the model with precise target descriptions and rich contextual knowledge, improving segmentation performance on emerging objects. Third, the Visual Prompt Enhancer (VPE) further supports the recognition of novel entities by incorporating internet-sourced reference images. To ensure efficiency, the WebSense module adaptively determines whether retrieval is necessary based on the relevance of user inputs. By leveraging the latest online multimodal information, ROSE enables effective segmentation of both novel and emerging objects in a resource-aware manner.

This work makes the following key contributions:

\begin{itemize}[topsep=0pt,itemsep=0pt,leftmargin=20pt]
\item We introduce the Novel Emerging Segmentation Task (NEST), which challenges models to segment (i) novel entities unrecognizable by MLLMs and (ii) emerging entities requiring real-time information retrieval. This capability is essential for developing robust intelligent perception systems that can effectively adapt to and comprehend continuously evolving environments.
\item We establish a \textit{\textbf{NEST}} benchmark for novel emerging segmentation.
Considering real-time data evolution, we develop an automated data engine that continuously constructs up-to-date datasets to evaluate models' novel emerging segmentation capabilities.
\item We propose \textit{\textbf{ROSE}}, a plug-and-play method that augments any MLLM-based segmentation model with the ability to segment novel and emerging entities. ROSE integrates four components to retrieve and employ up-to-date multimodal information from the web.

\item Experimental results shows that existing MLLM-based segmentation methods struggle to segment novel and emerging entities. Our proposed ROSE effectively addresses this limitation, outperforming a strong commercial retrieval baseline built on Gemini-2.0-Flash search by 19.2\% in gIoU.

\end{itemize}

\section{Related Work}
\textbf{Referring Expression Segmentation (RES).}
RES~\cite{ding2025multimodal,lisa,read,vlt,GREx} aims to segment target objects in images based on textual descriptions.
Early works~\cite{lstm-cnn,rmi,rrn} extract visual features using CNN and encode language expressions through LSTM. These extracted features are then fused through concatenation or other simple operations to create multi-modal representations.
VLT~\cite{vlt,vltpami} first introduces the transformer architecture, which reformulates the RES task as an attention problem and proposes to use language features to query the vision features, generating results by decoding the transformer response.
Liu~\etal~\cite{gres} proposes the Generalized Referring Expression Segmentation (GRES) task, supporting both the multi-target and empty-target scenarios.
LLMs/MLLMs~\cite{gpt4,llama3,llava,deepseek-vl,deepseek-v3,internvl} have revolutionized vision-language tasks by demonstrating remarkable capabilities in common-sense reasoning, opening up exciting new possibilities for RES~\cite{ding2025multimodal,MeViSv2}.
Building on these advances, LISA~\cite{lisa} innovatively introduces the [SEG] token, enabling the processing of expressions that require complex reasoning and common-sense knowledge.
SESAME~\cite{sesame} employs model chaining and joint training to tackle false premise failures in MLLMs.
READ~\cite{read} guides MLLMs on where to focus attention during interactive reasoning by treating similarity as reference points.

\noindent\textbf{Retrieval-Augmented Generation (RAG).}
RAG~\cite{yu2022generate,searchlvlms} has garnered significant attention in NLP~\cite{ragsurvey,realm,lazaridou2022internet} and multimodal communities~\cite{reveal,ra-cm3,searchlvlms}.
Early methods like REALM~\cite{realm} retrieve the top-k relevant snippets and use LLMs to generate k responses, combining them for QA tasks.
Yu~\etal~\cite{yu2022generate} employ search engines to enhance LLMs on zero-shot knowledge-intensive tasks.
Lazaridou~\etal~\cite{lazaridou2022internet} apply few-shot prompting to leverage Google search results for factual and up-to-date QA.
RA-CM3~\cite{ra-cm3} leverages external memory retrieval for text and image generation.
SearchLVLMs~\cite{searchlvlms} empowers MLLMs with real-time Internet search during inference.

\vspace{-3.2mm}
\section{NEST Dataset}

The Novel Emerging Segmentation Task requires models to segment objects by acquiring up-to-date information.
However, constructing a fixed, long-term dataset for evaluation is impractical, as such datasets may eventually be incorporated into the training of future MLLM-based segmentation models, introducing the risk of data leakage.
Moreover, the collection, creation, and segmentation annotation of such datasets are labor-intensive, making it infeasible to maintain comprehensive, continually updated benchmarks for newly emerging entities.
To address these challenges, we develop an automatic annotation pipeline that efficiently generates high-quality and continually up-to-date evaluation data for novel emerging segmentation task.
\begin{figure*}[!ht]
	\centering
 	\includegraphics[width=0.97\linewidth]{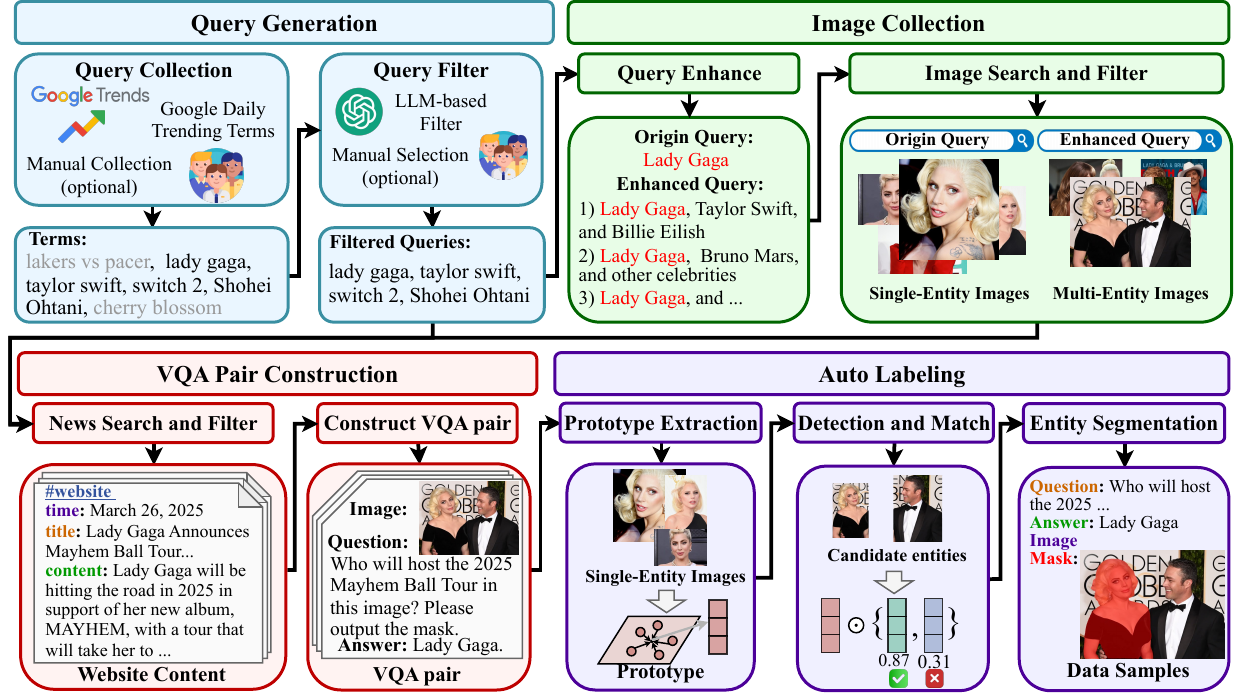}
        \vspace{-2mm}
	\caption{\textbf{NEST Data Engine.}
    We introduce an automated annotation pipeline that efficiently generates high-quality evaluation samples for the novel emerging segmentation task. The pipeline leverages time-specific queries to continuously collect news content and corresponding relevant images for constructing VQA pairs and automatically generating mask annotations, enabling a comprehensive and reliable evaluation of models' abilities to segment emerging entities.
    }\label{Fig:DataEngine}
    \vspace{-1\baselineskip}
\end{figure*}

\subsection{Automated NEST Data Engine}

Each data sample in NEST consists of questions~$\mathcal{Q}$, textual answers~$\mathcal{A}$, reference images~$\mathcal{I}_m$, and segmentation masks~$M$, providing a comprehensive foundation for evaluating models' novel emerging segmentation capabilities.

\noindent\textbf{Query Generation.} As shown in \cref{Fig:DataEngine}, we construct the query collection $Q$ primarily using Google Trends, supplemented with a small set of manually curated queries. Google Trends is a public platform that analyzes the popularity of search queries across regions and languages, surfacing trending keywords that reflect current user interests. While it offers valuable insights, the data is often noisy and skewed toward domains such as sports, entertainment, and politics. To improve domain diversity, we manually augment the query set with popular terms from technology, economics, and other underrepresented fields. However, many trending terms, \eg, abstract concepts like “Google stock”, are not directly associated with segmentable objects. To address this, we employ large language models (LLMs) such as LLaMA 3~\cite{llama3} and GPT-4o~\cite{gpt4o} to filter unqualified queries. This filtering process reveals that most viable segmentable queries correspond to people or products. The resulting filtered query set $\widetilde{Q}$ consists of concrete and clearly segmentable entities that maintain both topical relevance and category diversity, thereby enabling the effective construction of up-to-date segmentation samples.

\noindent\textbf{Image Collection.} We use the filtered query collection $\widetilde{Q}$ to retrieve corresponding images $\mathcal{I}$ from search engines for evaluating models' novel emerging segmentation capabilities. However, we observe that search engine results for many queries typically return images containing only a single visual entity, significantly reducing the task's complexity and failing to reflect realistic segmentation challenges. To address this limitation, we design an LLM-based query enhancement strategy that increases query complexity by encouraging the retrieval of images containing both the target entity and related entities. For example, instead of retrieving images featuring only “Mbappé”, the enhanced query yields images that also include “Neymar” and “Messi”, increasing visual ambiguity and segmentation difficulty. We submit both the original queries $\widetilde{Q}_o$ and the enhanced queries $\widetilde{Q}_e$ to search engines, resulting in two image sets: single-entity images $\mathcal{I}_s$ and multi-entity images $\mathcal{I}_m$. For $\mathcal{I}_s$, we apply CLIP-based feature clustering~\cite{clip} and retain only the images from the largest cluster to ensure consistency and visual quality. These filtered images are used to support our auto-labeling pipeline. For $\mathcal{I}_m$, we apply an object detector~\cite{yolov8} to filter out remaining single-entity samples. The resulting multi-entity images serve as the primary source for constructing the final segmentation dataset, ensuring higher task difficulty and diversity.

\noindent\textbf{VQA Pair Construction.} For each query retrieved from Google Trends, we also collect its associated metadata, including the query term, topic category, related keywords, and a set of related news articles. Each news item contains a link, publication timestamp, snippet, and other metadata. To reduce redundancy, we apply a temporal filtering strategy based on the observation that news articles published within a short time window often refer to the same event. Specifically, for articles published within a 3-day window, we retain only one representative article as the primary source for constructing visual question answering (VQA) pairs. We further refine this set by filtering out near-duplicate articles based on snippet similarity. To construct VQA samples, we employ LLMs to generate contextually relevant questions $\mathcal{Q}$ from the retained news content, where the original query $\widetilde{Q}_o$ serves as the answer $\mathcal{A}$. Our prompt design ensures that the generated questions are natural and do not explicitly mention the query term, thereby requiring genuine comprehension rather than simple keyword matching. Each resulting VQA sample is represented as a triplet $(\mathcal{Q}, \mathcal{A}, \mathcal{I}_m)$, where $\mathcal{Q}$ is the generated question, $\mathcal{A}$ is the answer, and $\mathcal{I}_m$ is the corresponding multi-entity image containing the target entity.

\begin{figure*}[!ht]
	\centering
 	\includegraphics[width=1\linewidth]{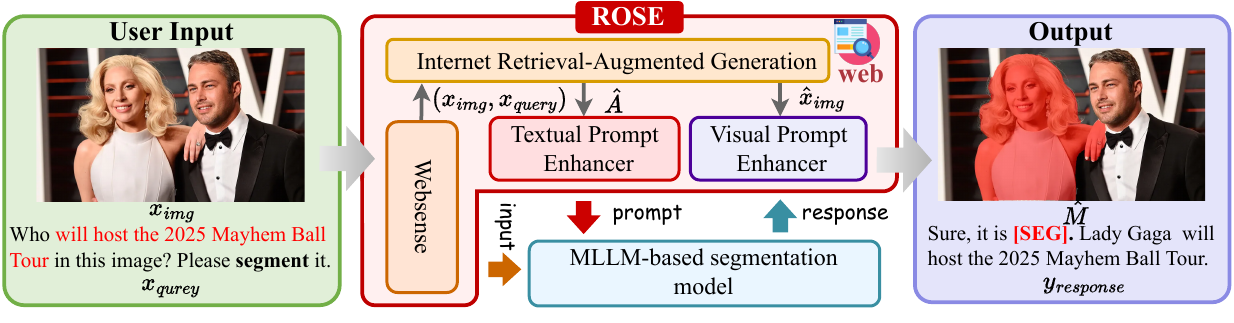}
        \vspace{-7.6mm}
	\captionof{figure}{\textbf{Architecture overview of ROSE.} Given a user input (image and question), ROSE first employs the WebSense module to determine whether internet retrieval is needed. If so, the Internet Retrieval-Augmented Generation module retrieves relevant textual and visual data from the web. The retrieved content is then processed by the Textual and Visual Prompt Enhancer to generate enriched prompts for the MLLM-based segmentation model, which ultimately produces accurate segmentation masks for novel and emerging entities.
    }
    \label{fig:rose}
\end{figure*}
\noindent\textbf{Auto Labeling.} We employ an automated labeling pipeline to generate segmentation masks $M$ for target entities, using the single-entity image set $\mathcal{I}_s$ associated with the ground-truth query $\widetilde{Q}_o$. First, we extract CLIP features $f_s$ from the clustered single-entity images $I_s$ to serve as the target entity representation. Next, we apply an object detector to the corresponding multi-entity images $\mathcal{I}_m$, identifying a set of entity proposals $\{E_i\}_{i=1}^n$. For each detected entity $E_i$, we crop its bounding region and extract CLIP features $f_i$. We then compute the cosine similarity between each $f_i$ and the target representation $f_s$, selecting the entity with the highest similarity score. If the similarity exceeds a predefined threshold $\tau$, the corresponding entity is selected as the target. We then pass its bounding box coordinates to the SAM mask decoder~\cite{sam} to generate a high-quality segmentation mask $M$. This entire process is fully automated and requires no manual human intervention, greatly reducing the annotation burden while enabling scalable generation of up-to-date segmentation evaluation data.

\subsection{NEST Dataset Analysis}

We leverage our data engine to systematically collect and process web data by issuing queries to scrape news content between March 23, 2025 and April 11, 2025. The resulting \textit{NEST} dataset contains 1,548 evaluation samples, primarily focusing on people and products across diverse domains including economics, technology, politics, entertainment, sports, and society. On average, each image contains 2.7 valid entities, increasing task complexity and mitigating hallucinations from large language models. Additionally, each image is paired with an average of 1.6 unique questions, enriching the diversity of query formulations. Further details are provided in Supplementary Materials.

\section{Method}
\subsection{Architecture Overview.} 
The architecture of ROSE is shown in Fig.~\ref{fig:rose}. 
Given an input image $x_{img}$ and a user query $x_{query}$, ROSE first employs the WebSense module to determine whether internet retrieval is necessary. 
If needed, the IRAG module leverages $x_{img}$ and $x_{query}$ to retrieve answers $\hat{A}$ and relevant image~$\hat{x}_{img}$ information from the internet.
Then, the TPE module leverages $\hat{A}$ to generate an enhanced textual prompt $P = f(x_{query}, \hat{A}, K_{ext})$ to improve the MLLM-based segmentation model's ability to segment emerging entities, where $K_{ext}$ represents extra background knowledge. 
For novel entities that may still be challenging to segment, the VPE module leverages the retrieved images $\hat{x}_{img}$ to extract their prototype features~$\textbf{f}_s$, which are then used to assist the model in segmenting novel entities.
This comprehensive and unified method enables ROSE to effectively segment both novel entities that are entirely absent from MLLMs' training data and emerging entities that require continuously up-to-date information, thus expanding the segmentation capability from the original model's knowledge space $K$ to an enhanced space $S=K \cup E$, where $E$ represents the external knowledge database.

\subsection{Internet Retrieval-Augmented Generation.}
\label{sec:irag}
To retrieve up-to-date visual and textual information from the internet for enhanced segmentation, we develop an Internet Retrieval-Augmented Generation~(IRAG) module based on LangChain~\cite{langchain} framework.
This module processes user query~$x_{query}$ by first generating optimized search queries~$q$ using large language models~(LLMs), then retrieving relevant content through search engines. 
The retrieved content is split into manageable chunks $\{C_i\}_{i=1}^n$, vectorized using embedding techniques $\mathbf{E}(C_i) \in \mathbb{R}^d$, and stored in a vector database $\mathcal{D} = \{(\mathbf{E}(C_i), C_i)\}_{i=1}^n$ for efficient retrieval. 
A map-reduce method with specialized prompts processes these chunks to extract the most relevant information, which is then synthesized into an answer candidate summary containing the potential answers $\{A_j\}_{j=1}^m$ to the user's query.
However, since a question's answer is not always unique, the answer candidate summary may contain multiple valid answers.
Therefore, we need to leverage the user-provided image $x_{img}$ to narrow down the range of potential answers.
To accomplish this, we employ Google Cloud Vision to analyze and extract multiple entities $\{E_k\}_{k=1}^l$ from the input image $x_{img}$. 
We choose not to use MLLMs for entity extraction because they lack the ability to accurately identify novel entities.
Then, we compare the entities in the answer candidate summary $\{A_j\}_{j=1}^m$ with the detected entities $\{E_k\}_{k=1}^l$ to determine the correct answer $\hat{A}$. 
If no matching entities are found in the image, we select the entity with the highest confidence score from the answer candidate summary as the correct answer.
Finally, we use $\hat{A}$ as a keyword to retrieve relevant images~$\hat{x}_{img}$ from the internet.

\subsection{Textual Prompt Enhancer.}
\label{sec:tpe}
To effectively leverage the retrieved answer~$\hat{A}$ from IRAG, we design a textual prompt enhancer (TPE) to generate an enhanced textual prompt that improves the model's understanding of the target entity.
This module integrates the query~$x_{query}$, the answer~$\hat{A}$ retrieved from the IRAG module, and the answer's extra background knowledge~$K_{ext}$ to create an optimized prompt. The background knowledge~$K_{ext}$ is obtained by retrieving the target's introduction from the Internet using~$\hat{A}$ as the search term.
By strategically combining the answer information from the IRAG module with comprehensive background knowledge, the module produces prompts that are both informative and directive, enabling the MLLM-based segmentation model to accurately identify and segment the target object within the image.

\subsection{Visual Prompt Enhancer.}
\label{sec:vpe}
VPE enhances the model's capability to segment novel entities by verifying and correcting results when the MLLM-based segmentation model fails to produce accurate segmentation.
Using images~$\hat{x}_{img}$ retrieved from the internet, we perform clustering and retain only the largest cluster. We then extract CLIP~\cite{clip} features from these images to obtain a prototype feature representation $\mathbf{f}_s$.
We compare this prototype feature with CLIP features extracted from the foreground region segmented by the MLLM-based segmentation model. 
A low similarity score indicates that the MLLM has failed to correctly identify the target entity. 
In such cases, we employ an object detector to identify multiple entities $\{E_i\}_{i=1}^n$ within the input image $x_{img}$. 
For each detected entity $E_i$, we extract CLIP features $\mathbf{f}_{i}$ from the cropped region. 
By computing the similarity between each $\mathbf{f}_{i}$ and the prototype feature $\mathbf{f}_s$, we identify the entity with the highest similarity score.
When this score exceeds a predefined confidence threshold $\tau$, we designate the corresponding entity as the target. 
Finally, we use the bounding box coordinates of the identified target as input to SAM's mask decoder~\cite{sam}, thereby generating a high-quality segmentation mask $\hat{M}$ for the target entity.

\subsection{WebSense.}
\label{sec:websense}
To reduce computational and network resources, the WebSense module is designed to intelligently determine whether the IRAG module should be activated based on $x_{query}$.
While certain queries require real-time external knowledge retrieval to ensure accurate segmentation, others can be addressed using the internal knowledge of existing MLLMs without accessing external resources. This selective activation mechanism improves computational efficiency and reduces latency by invoking retrieval only when necessary.
WebSense adopts a two-tier decision architecture. 
In the first stage, a lightweight rule-based filter quickly screens queries using predefined heuristics, \eg, time-sensitive rules. For queries that are ambiguous or semantically complex, a large language model~\cite{llama3} is employed to perform deeper semantic analysis and classify whether retrieval is needed.

\section{Experiments}
\label{sec:experiment}
\textbf{Implementation Details.}
Our method is primarily built on the LangChain~\cite{langchain} framework, which enables efficient retrieval and reasoning capabilities.
We employ Llama-3-8B~\cite{llama3} as our foundation language model, with a knowledge cutoff date of December 2023, which ensures no knowledge leakage for the NEST dataset.
For prototype feature extraction, we employ CLIP-ViT-L/32~\cite{clip} to obtain rich visual representations. 
Additionally, we use YOLOv8~\cite{yolov8} as our object detector to identify potential regions of interest in the input images. 
All evaluations are conducted with a single NVIDIA 48G A6000 GPU.

\noindent\textbf{Datasets.} We evaluate the novel emerging segmentation capabilities of MLLM-based segmentation models using two datasets: NEST and NEST+. To enable more comprehensive and realistic evaluation, we construct NEST+ by combining NEST with ReasonSeg~\cite{lisa}, RefCOCO, RefCOCO+\cite{referit}, and RefCOCOg\cite{refcoco}. The composite dataset NEST+ supports joint evaluation across novel emerging segmentation, reasoning-based segmentation, and traditional referring segmentation tasks.

\noindent\textbf{Baselines.} We conduct experiments on several state-of-the-art MLLM-based segmentation models~\cite{lisa,sesame,read}. However, these models inherently lack internet retrieval functionality, limiting their ability to handle novel or time-sensitive queries. To ensure fair comparisons, we implement enhanced baselines by integrating commercial retrieval-augmented models, GPT-4o mini Search~\cite{gpt4o} and Gemini 2.0 Flash Search~\cite{gemini}, which are equipped with built-in internet search capabilities.

\noindent\textbf{Evaluation Metrics.}
Following~\cite{vlt,lisa}, we adopt two standard segmentation metrics: gIoU and cIoU.
Additionally, we use Acc.~to comprehensively evaluate the question-answering capability of RAG models.

\subsection{Comparison with the State-of-the-Art Methods}
\begin{table*}[!t]
    \centering
    \footnotesize
    \caption{\textbf{Novel Emerging Segmentation results on NEST dataset.} Furthermore, we partition NEST into a novel entity split and an emerging entity split based on LLaVA-v1.5-7B's~\cite{llava} knowledge, which is the foundation model employed by LISA-7B~\cite{lisa}, SESAME-7B~\cite{sesame}, and READ-7B~\cite{read}.}
    \vspace{-2.5mm}
    \setlength{\tabcolsep}{12.6pt}
    \renewcommand{\arraystretch}{1.08}
    \begin{tabular}{l|c|c|cc|cc|cc}
        \hline
    {\textbf{Method}} & \textbf{RAG} & \textbf{Acc.} & \multicolumn{2}{c|}{\textbf{Novel Entity}} & \multicolumn{2}{c|}{\textbf{Emerging Entity}} & \multicolumn{2}{c}{\textbf{Overall}} \\
    & & & gIoU  & cIoU & gIoU  & cIoU & gIoU  & cIoU  \\
    \hline
    \rowcolor{myrowgray!50}
    CRIS~\cite{cris} & \xmark & - & 36.8 & 27.3 & 40.5 & 30.3 & 38.9 & 29.1 \\
    
    GRES~\cite{gres} & \xmark & - & 37.8 & 36.6 & 44.2 & 36.1 & 41.4 & 36.3 \\
    \rowcolor{myrowgray!50}
    Grounded-SAM~\cite{grounded-sam} & \xmark & - & 39.0 & 31.7 & 53.8 & 40.3 & 47.4 & 36.7 \\
    SEEM~\cite{seem} & \xmark & - & 41.8 & 27.4 & 47.1 & 38.2 & 44.8 & 33.7 \\
    \hline
    \rowcolor{myrowgray!50}
    LISA-7B~\cite{lisa} & \xmark & - & 38.4 & 28.5 &56.5 & 47.5  & 48.7 & 39.3 \\
    
    SESAME-7B~\cite{sesame} & \xmark & - & 11.1 & \ \ 7.9 & 14.5 & 14.0 & 13.1 & 11.6 \\
    \rowcolor{myrowgray!50}
    READ-7B~\cite{read} & \xmark & - & 19.2 & 17.6 & 25.1 & 22.0 & 22.5 & 20.2 \\
    \hline
    LISA-7B+ GPT-4o mini Search~\cite{gpt4o} & \cmark & 68.1 & 35.4 & 30.8 & 67.0 & 63.1 & 53.5 & 49.0 \\
    \rowcolor{myrowgray!50}
    SESAME-7B+ GPT-4o mini Search & \cmark & 67.8 & 27.8 & 22.6 & 42.3 & 41.7 & 36.1 & 33.8 \\
    
    READ-7B+ GPT-4o mini Search & \cmark & 68.7 & 34.9 & 31.7 & 47.5 & 43.0 & 42.1 & 38.2 \\
    \rowcolor{myrowgray!50}
    LISA-7B+ Gemini-2.0 Flash Search~\cite{gemini} & \cmark & 69.6 & 35.2 & 29.6& 67.8 & 65.3 & 53.8 & 49.3 \\
    
    SESAME-7B+ Gemini-2.0 Flash Search & \cmark & 71.1 & 30.3 & 23.8 & 46.0 & 45.6 & 39.2 & 36.6 \\
    \rowcolor{myrowgray!50}
    READ-7B+ Gemini-2.0 Flash Search & \cmark & 70.0 & 37.1 & 31.6 & 50.2 & 44.9 & 44.6 & 39.3 \\
    \hline
    \rowcolor{cyan!10}   \textbf{LISA-7B + ROSE~(ours)} & \cmark & \underline{73.4} & \underline{67.0} & \textbf{65.7} & \textbf{77.5} & \underline{70.7} & \textbf{73.0} & \textbf{68.6} \\
  
    SESAME-7B + ROSE~(ours) & \cmark & 72.9 & 65.9 & 62.5 & 74.1 & 70.5 & 70.6 & 67.2 \\
       \rowcolor{myrowgray!50}
    READ-7B + ROSE~(ours) & \cmark & \textbf{74.2} & \textbf{67.1} & \underline{63.4} & \underline{76.0} & \textbf{71.7} & \underline{72.2} & \underline{68.3} \\
    \bottomrule
    \end{tabular}
    \label{tab:nest}
\end{table*}

\begin{table*}[!t]
    \centering
    \footnotesize
    \caption{\textbf{Mixed Dataset NEST+ Results.} ReasonSeg is sourced from~\cite{lisa}, while RefSeg is derived from RefCOCO, RefCOCO+~\cite{referit} and RefCOCOg~\cite{refcoco} datasets. }
       \vspace{-2.5mm}
    \setlength{\tabcolsep}{12.6pt}
    \renewcommand{\arraystretch}{1.08}
    \begin{tabular}{l|c|cc|cc|cc|cc}
        \hline
    {\textbf{Method}} & \textbf{RAG} & \multicolumn{2}{c|}{\textbf{NEST}} & \multicolumn{2}{c|}{\textbf{ReasonSeg}} & \multicolumn{2}{c|}{\textbf{RefSeg}} & \multicolumn{2}{c}{\textbf{Overall}} \\
    & & gIoU  & cIoU & gIoU  & cIoU & gIoU  & cIoU & gIoU  & cIoU \\
    \hline
    \rowcolor{myrowgray!50}
    CRIS~\cite{cris} & \xmarkg & 39.6 & 29.2 & 18.2 & 19.7 & 56.0 & 50.4 & 40.5 & 26.6 \\
    
    GRES~\cite{gres} & \xmarkg & 42.7 & 35.6 & 20.3 & 17.7 & 60.8 & 56.2 & 43.8 & 30.0 \\
    \rowcolor{myrowgray!50}
    Grounded-SAM~\cite{grounded-sam} & \xmarkg & 47.8 & 34.4 & 20.6 & 15.6 & 42.4 & 31.4 & 43.8 & 27.7 \\
    SEEM~\cite{seem} & \xmarkg & 47.3 & 36.6 & 33.2 & 25.4 & 21.6 & 17.3 & 41.7 &29.7\\
    \hline
    \rowcolor{myrowgray!50}
    LISA-7B~\cite{lisa} & \xmarkg & 51.1 & 39.0 & 42.5 & 44.2 & 54.9 & 53.7 & 50.9 & 40.6 \\
    
    SESAME-7B~\cite{sesame} & \xmarkg & 14.1 & 11.2 & 30.9 & 28.7 & 62.7 & \underline{61.7}& 25.5 & 17.8 \\
    \rowcolor{myrowgray!50}
    READ-7B~\cite{read} & \xmarkg & 22.3 & 18.3 & \underline{49.7} & \underline{54.2} & 63.9 & 59.9 & 33.4 & 28.3 \\
    \hline
\rowcolor{myrowgray!50}
    LISA-7B + ROSE~(ours) & \cmark & \textbf{75.3} & \textbf{67.4} & 42.2 & 44.1 & 54.4 & 52.4 & \underline{67.6} & \underline{60.7} \\

    SESAME-7B + ROSE~(ours) & \cmark & 70.1 & 63.5 & 33.7 & 31.6 & \textbf{68.4} & \textbf{66.3} & 65.8 & 54.2 \\
    \rowcolor{cyan!10}
    \textbf{READ-7B + ROSE~(ours)} & \cmark & \underline{71.6} & 
    \underline{65.2} & \textbf{50.3} & \textbf{54.9} & \underline{64.7} & 60.9 & \textbf{67.9} & \textbf{62.4} \\
    \bottomrule
    \end{tabular}
    \vspace{-0.2cm}
    \label{tab:nest+}
\end{table*}

\noindent\textbf{Experiments on NEST Dataset.}
The novel emerging segmentation results on the NEST dataset are shown in \Cref{tab:nest}.
It is worth noting that existing works fail to handle this task,
while our method successfully accomplishes it by retrieving up-to-date information, achieving more than 30\% gIoU performance improvement.
Unlike traditional referring segmentation, the novel emerging segmentation task presents a unique challenge by requiring models to identify and segment entities that have emerged after their training cutoff date.
Only by leveraging the user's multimodal input to retrieve information from the internet can the model perform well on this task.
Existing works have no effective way to identify entities beyond their knowledge cutoff, but our model explicitly exploits Retrieval-Augmented Generation (RAG) to more effectively address this challenge.

Furthermore, we also compare our method with vanilla two-stage baselines combining MLLM-based segmentation models with GPT-4o mini Search~\cite{gpt4o} and Gemini-2.0 Flash Search~\cite{gemini}, which are advanced commercial LLMs with built-in internet retrieval capabilities.
This baseline first uses them to generate an answer based on the input question, then employs MLLM-based segmentation models~\cite{lisa,sesame,read} to produce the segmentation mask.
For the intermediate prompt to MLLM-based segmentation models, we use the template ``\textit{Please segment \{answer\} in this image.}'' where \textit{{answer}} is the response from the commercial LLM with built-in internet retrieval capabilities.
As shown in \Cref{tab:nest}, ROSE significantly outperforms this two-stage method for two main reasons:
(1) Our TPE module provides more accurate target information and richer background knowledge, enhancing the MLLM's understanding of the target objects,
and (2) our model leverages internet-sourced images to support novel entity segmentation
, effectively addressing the novel entity segmentation problem.
Additional experiments and detailed analysis are provided in Supplementary Materials.

\noindent\textbf{Experiments on Mixed Dataset NEST+.}
To further evaluate the generalization ability of our method, we conduct experiments on a mixed dataset NEST+.
The NEST+ simulates real-world scenarios involving real-time retrieval, reasoning, and traditional referring segmentation.
As shown in \Cref{tab:nest+}, ROSE significantly enhances performance on the NEST split while maintaining competitive results on both the ReasonSeg and RefSeg splits.
This demonstrates that ROSE improves the ability of novel emerging segmentation while maintaining performance on traditional tasks.

\subsection{Ablation Study}
Here, we perform an ablation study with LISA-7B~\cite{lisa} as the baseline.
To ensure that each evaluation requires retrieval, the WebSense module is excluded from our ablation studies, allowing for a more comprehensive assessment of the model's ability to segment novel and emerging entities.

\begin{figure*}[!t]
\vspace{-6.6mm}
	\centering
 	\includegraphics[width=0.98\linewidth]{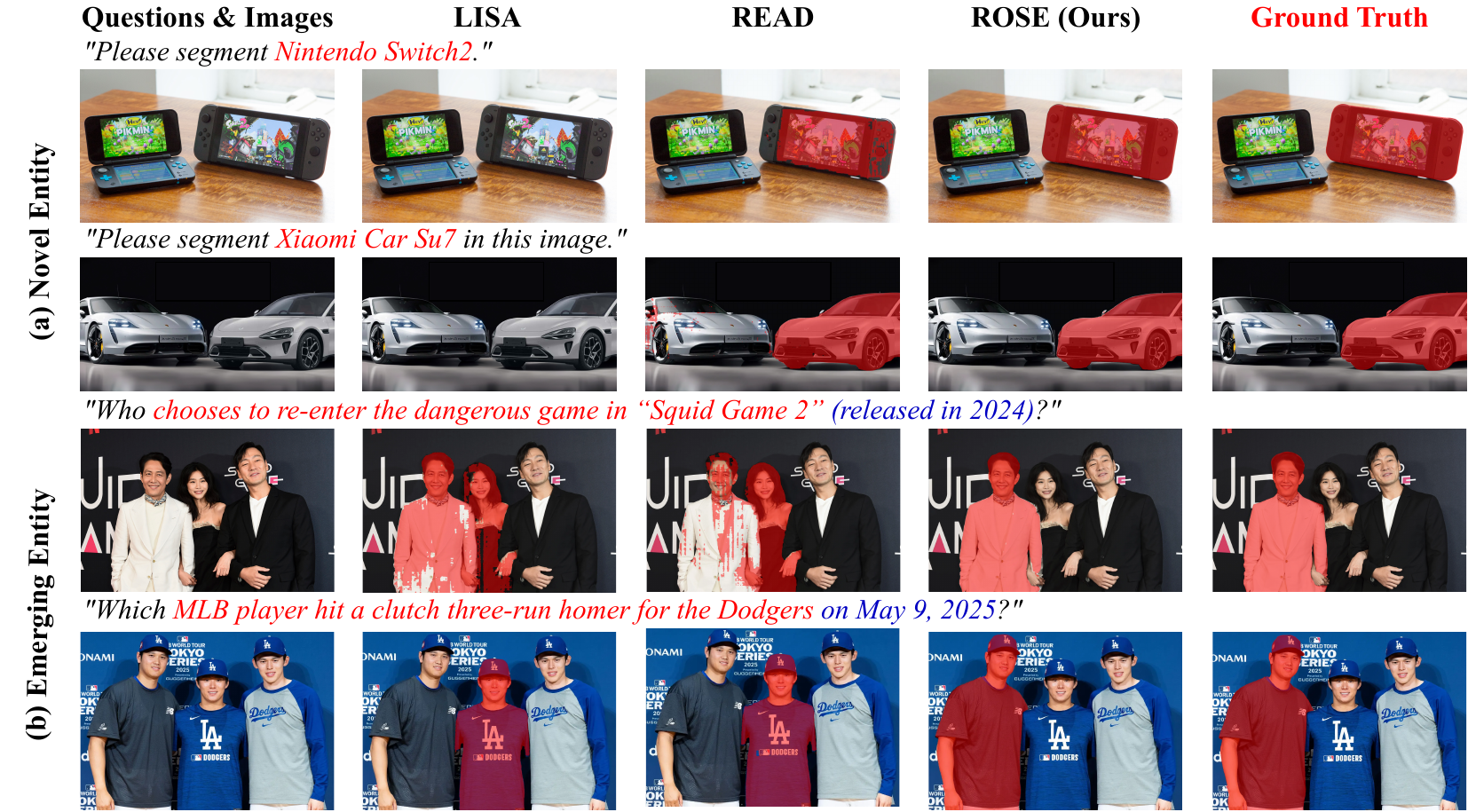}
        
	\caption{\textbf{Qualitative results} comparing LISA~\cite{lisa}, READ~\cite{read}, and ROSE on novel and emerging entities. ROSE accurately segments unseen and newly emerging targets, while LISA struggles due to a lack of up-to-date knowledge or inability to recognize new entities.
    }
    \label{fig:visualization}
\end{figure*}

\begin{table}
    \begin{center}
    \footnotesize
    \setlength{\tabcolsep}{1.3mm}
    \renewcommand{\arraystretch}{1.08}
    \caption{\textbf{Ablation Study.}  Impact of different ROSE components on novel emerging segmentation, conducted on the NEST dataset. The data split follows the same approach as in~\cref{tab:nest}.}
       \vspace{-2.5mm}
    \begin{tabular}{l|cc|cc|cc}
    \hline
    \textbf{Method} & \multicolumn{2}{c|}{\textbf{Novel Entity}} & \multicolumn{2}{c|}{\textbf{Emerging Entity}} & \multicolumn{2}{c}{\textbf{Overall}} \\
    & gIoU  & cIoU  & gIoU  & cIoU  & gIoU  & cIoU  \\
    \hline
    \rowcolor{myrowgray!50}
    LISA-7B~\cite{lisa} & 38.4 & 28.5 & 56.5 & 47.5 & 48.7 & 39.3 \\
    +ROSE~(IRAG only) & 40.4 & 30.9 & 67.1 & 62.7 & 55.7 & 49.1 \\
    \rowcolor{myrowgray!50}
    +ROSE~(IRAG+TPE) & 41.3 & 31.6 & \underline{73.3} & \underline{67.2} & 59.6 & 51.8 \\
    +ROSE~(IRAG+VPE) & \underline{64.9} & \underline{61.7} & 71.7 & 65.5 & \underline{68.7} & \underline{63.8} \\
    \rowcolor{cyan!10}
    \textbf{+ROSE~(Full)} & \textbf{68.6} & \textbf{67.2} & \textbf{79.4} & \textbf{72.3} & \textbf{74.7} & \textbf{70.1} \\
    \bottomrule
    \end{tabular}
    \label{tab:ablation}
    \end{center}
    \vspace{-9.5mm}
\end{table}

\noindent\textbf{Effect of Internet Retrieval-Augmented Generation Module (IRAG).}
As shown in Table~\ref{tab:ablation}, adding the IRAG module brings effective improvements over the baseline method LISA-7B, with overall gIoU increasing by 7.0\% and cIoU increasing. This substantial enhancement confirms that retrieving up-to-date information from the internet is crucial for novel emerging segmentation, as the model cannot rely solely on pre-trained knowledge to identify and segment novel or emerging entities.

\noindent\textbf{Effect of Textual Prompt Enhancer (TPE).}
As shown in \Cref{tab:ablation}, comparing IRAG-only with IRAG+TPE, we observe that TPE improves performance on emerging entities (gIoU +6.2\% and cIoU +4.5\%). This demonstrates that well-structured textual prompts effectively integrate retrieved knowledge with the original instruction.

\noindent\textbf{Effect of Visual Prompt Enhancer (VPE).}
As shown in \Cref{tab:ablation}, comparing IRAG-only with IRAG+VPE, we observe that VPE significantly improves performance on novel entities (cIoU +24.5\%) and overall performance (gIoU +13.0\%). This indicates that enhancing the model's visual understanding through retrieved images significantly improves its ability to segment novel entities by providing visual references that complement the textual information.

\subsection{Qualitative Results}
\cref{fig:visualization} presents qualitative results demonstrating the effectiveness of ROSE. We present four examples comparing ROSE with LISA~\cite{lisa} and READ~\cite{read} on two key challenges in novel emerging segmentation: novel entities and emerging entities. For novel entities in the first two rows, ROSE accurately segments the referred targets, while LISA fails due to its inability to recognize unseen entities. For example, in the 2nd row, LISA produces no output for the novel entity Xiaomi SU7, whereas ROSE correctly segments it. For emerging entities in the last two rows, LISA similarly struggles due to outdated knowledge. In the 4th row, given the question “\textit{Which MLB player hit a clutch three-run homer for the Dodgers on May 9, 2025? }”, LISA segments the wrong person, while ROSE correctly identifies and segments the actual player. More visualization results are provided in the Supplementary Materials.

\vspace{-1.5mm}
\section{Conclusion}
We introduce the novel emerging segmentation task (NEST), which requires segmenting (i) novel entities absent from MLLMs' training data and (ii) emerging entities that demand up-to-date external knowledge. To support this, we construct the {NEST} dataset using an automated pipeline that collects real-time image–news pairs. We propose ROSE, a plug-and-play approach that enhances MLLM-based segmentation models through real-time internet retrieval. Extensive experiments show that ROSE significantly improves performance on novel emerging segmentation while maintaining competitive results on standard benchmarks. Our work highlights the potential of combining retrieval with multimodal models for real-time understanding of newly emerging entities.

\clearpage
\footnotesize{\paragraph{Acknowledgement.} This work was supported by the Science and Technology Commission of Shanghai Municipality (No.~25511103600) and the National Natural Science Foundation of China (NSFC) under Grant No. 62472104.


\appendix

{
    \small
    \bibliographystyle{ieeenat_fullname}
    \bibliography{main}

@String(PAMI = {IEEE Trans. Pattern Anal. Mach. Intell.})

@String(IJCV = {Int. J. Comput. Vis.})

@String(CVPR= {IEEE Conf. Comput. Vis. Pattern Recog.})

@String(ICCV= {Int. Conf. Comput. Vis.})

@String(ECCV= {Eur. Conf. Comput. Vis.})

@String(ICLR = {Int. Conf. Learn. Represent.})

@String(PAMI  = {IEEE TPAMI})

@String(IJCV  = {IJCV})

@String(CVPR  = {CVPR})

@String(ICCV  = {ICCV})

@String(ECCV  = {ECCV})

@String(NeurIPS  = {NeurIPS})

@String(ICLR  = {ICLR})

@String(EMNLP = {EMNLP})

@String(ICML = {ICML})

@inproceedings{lisa,
  title={Lisa: Reasoning segmentation via large language model},
  author={Lai, Xin and Tian, Zhuotao and Chen, Yukang and Li, Yanwei and Yuan, Yuhui and Liu, Shu and Jia, Jiaya},
  booktitle=CVPR,
  xxxxpages={9579--9589},
  year={2024}
}

@inproceedings{gres,
  author       = {Chang Liu and
                  Henghui Ding and
                  Xudong Jiang},
  title        = {{GRES:} Generalized Referring Expression Segmentation},
  booktitle    = CVPR,
  xxxxpages        = {23592--23601},
  year         = {2023},
}

@article{llama3,
  title={The llama 3 herd of models},
  author={Grattafiori, Aaron and Dubey, Abhimanyu and Jauhri, Abhinav and Pandey, Abhinav and Kadian, Abhishek and Al-Dahle, Ahmad and Letman, Aiesha and Mathur, Akhil and Schelten, Alan and Vaughan, Alex and others},
  journal={arXiv preprint arXiv:2407.21783},
  year={2024}
}

@inproceedings{llava,
  title={Visual instruction tuning},
  author={Liu, Haotian and Li, Chunyuan and Wu, Qingyang and Lee, Yong Jae},
  booktitle=NeurIPS,
  year={2024}
}

@inproceedings{internvl,
  title={Internvl: Scaling up vision foundation models and aligning for generic visual-linguistic tasks},
  author={Chen, Zhe and Wu, Jiannan and Wang, Wenhai and Su, Weijie and Chen, Guo and Xing, Sen and Zhong, Muyan and Zhang, Qinglong and Zhu, Xizhou and Lu, Lewei and others},
  booktitle=CVPR,
  xxxxpages={24185--24198},
  year={2024}
}

@article{gpt4,
  title={Gpt-4 technical report},
  author={Achiam, Josh and Adler, Steven and Agarwal, Sandhini and Ahmad, Lama and Akkaya, Ilge and Aleman, Florencia Leoni and Almeida, Diogo and Altenschmidt, Janko and Altman, Sam and Anadkat, Shyamal and others},
  journal={arXiv preprint arXiv:2303.08774},
  year={2023}
}

@article{deepseek-vl,
  title={Deepseek-vl: towards real-world vision-language understanding},
  author={Lu, Haoyu and Liu, Wen and Zhang, Bo and Wang, Bingxuan and Dong, Kai and Liu, Bo and Sun, Jingxiang and Ren, Tongzheng and Li, Zhuoshu and Yang, Hao and others},
  journal={arXiv preprint arXiv:2403.05525},
  year={2024}
}

@article{deepseek-v3,
  title={Deepseek-v3 technical report},
  author={Liu, Aixin and Feng, Bei and Xue, Bing and Wang, Bingxuan and Wu, Bochao and Lu, Chengda and Zhao, Chenggang and Deng, Chengqi and Zhang, Chenyu and Ruan, Chong and others},
  journal={arXiv preprint arXiv:2412.19437},
  year={2024}
}

@article{seem,
  title={Segment everything everywhere all at once},
  author={Zou, Xueyan and Yang, Jianwei and Zhang, Hao and Li, Feng and Li, Linjie and Wang, Jianfeng and Wang, Lijuan and Gao, Jianfeng and Lee, Yong Jae},
  journal=NeurIPS,
  volume={36},
  year={2024}
}

@article{vltpami,
  title={{VLT}: Vision-language transformer and query generation for referring segmentation},
  author={Ding, Henghui and Liu, Chang and Wang, Suchen and Jiang, Xudong},
  journal=PAMI,
  volume={45},
  number={6},
  xxxxpages={7900--7916},
  year={2023},
  publisher={IEEE}
}

@inproceedings{vlt,
  title={Vision-language transformer and query generation for referring segmentation},
  author={Ding, Henghui and Liu, Chang and Wang, Suchen and Jiang, Xudong},
  booktitle=ICCV,
  xxxxpages={16321--16330},
  year={2021}
}

@inproceedings{lstm-cnn,
  title={Segmentation from natural language expressions},
  author={Hu, Ronghang and Rohrbach, Marcus and Darrell, Trevor},
  booktitle=ECCV,
  xxxxpages={108--124},
  year={2016}
}

@inproceedings{rmi,
  title={Recurrent multimodal interaction for referring image segmentation},
  author={Liu, Chenxi and Lin, Zhe and Shen, Xiaohui and Yang, Jimei and Lu, Xin and Yuille, Alan},
  booktitle=ICCV,
  xxxxpages={1271--1280},
  year={2017}
}

@inproceedings{rrn,
  title={Referring image segmentation via recurrent refinement networks},
  author={Li, Ruiyu and Li, Kaican and Kuo, Yi-Chun and Shu, Michelle and Qi, Xiaojuan and Shen, Xiaoyong and Jia, Jiaya},
  booktitle=CVPR,
  xxxxpages={5745--5753},
  year={2018}
}

@inproceedings{searchlvlms,
  title={SearchLVLMs: A Plug-and-Play Framework for Augmenting Large Vision-Language Models by Searching Up-to-Date Internet Knowledge},
  author={Li, Chuanhao and Li, Zhen and Jing, Chenchen and Liu, Shuo and Shao, Wenqi and Wu, Yuwei and Luo, Ping and Qiao, Yu and Zhang, Kaipeng},
  booktitle=NeurIPS,
  year={2024}
}

@inproceedings{cris,
  title={Cris: Clip-driven referring image segmentation},
  author={Wang, Zhaoqing and Lu, Yu and Li, Qiang and Tao, Xunqiang and Guo, Yandong and Gong, Mingming and Liu, Tongliang},
  booktitle=CVPR,
  xxxxxxxxpages={11686--11695},
  year={2022}
}

@article{grounded-sam,
  title={Grounded sam: Assembling open-world models for diverse visual tasks},
  author={Ren, Tianhe and Liu, Shilong and Zeng, Ailing and Lin, Jing and Li, Kunchang and Cao, He and Chen, Jiayu and Huang, Xinyu and Chen, Yukang and Yan, Feng and others},
  journal={arXiv preprint arXiv:2401.14159},
  year={2024}
}

@article{gpt4o,
  title={Gpt-4o system card},
  author={Hurst, Aaron and Lerer, Adam and Goucher, Adam P and Perelman, Adam and Ramesh, Aditya and Clark, Aidan and Ostrow, AJ and Welihinda, Akila and Hayes, Alan and Radford, Alec and others},
  journal={arXiv preprint arXiv:2410.21276},
  year={2024}
}

@article{gpt4v,
  title={Sparks of artificial general intelligence: Early experiments with gpt-4.},
  author={Bubeck, S{\'e}bastien and Chandrasekaran, Varun and Eldan, Ronen and Gehrke, Johannes and Horvitz, Eric and Kamar, Ece and Lee, Peter and Lee, Yin Tat and Li, Yuanzhi and Lundberg, Scott and others},
  journal={arXiv preprint arXiv:2303.12712},
  year={2023}
}

@article{gemini,
  title={Gemini: a family of highly capable multimodal models},
  author={Team, Gemini and Anil, Rohan and Borgeaud, Sebastian and Alayrac, Jean-Baptiste and Yu, Jiahui and Soricut, Radu and Schalkwyk, Johan and Dai, Andrew M and Hauth, Anja and Millican, Katie and others},
  journal={arXiv preprint arXiv:2312.11805},
  year={2023}
}

@article{langchain,
  author       = {Harrison Chase},
  title        = {LangChain: Building applications with large language models},
  year         = {2022},
  publisher    = {GitHub},
}

@inproceedings{read,
  title={Reasoning to Attend: Try to Understand How \textless\textsc{seg}\textgreater{} Token Works},
  author={Qian, Rui and Yin, Xin and Dou, Dejing},
  booktitle=CVPR,
  year={2025}
}

@article{MeViSv2,
  title={{MeViS}: A Multi-Modal Dataset for Referring Motion Expression Video Segmentation},
  author={Ding, Henghui and Liu, Chang and He, Shuting and Ying, Kaining and Jiang, Xudong and Loy, Chen Change and Jiang, Yu-Gang},
  journal=PAMI,
  year={2025},
  publisher={IEEE}
}

@article{GREx,
  title={{GREx}: Generalized Referring Expression Segmentation, Comprehension, and Generation},
  author={Ding, Henghui and Liu, Chang and He, Shuting and Jiang, Xudong and Jiang, Yu-Gang},
  journal=IJCV,
  year={2026},
  publisher={Springer}
}

@inproceedings{ra-cm3,
  title={Retrieval-augmented multimodal language modeling},
  author={Yasunaga, Michihiro and Aghajanyan, Armen and Shi, Weijia and James, Rich and Leskovec, Jure and Liang, Percy and Lewis, Mike and Zettlemoyer, Luke and Yih, Wen-tau},
  booktitle=ICML,
  year={2022}
}

@inproceedings{reveal,
  title={Reveal: Retrieval-augmented visual-language pre-training with multi-source multimodal knowledge memory},
  author={Hu, Ziniu and Iscen, Ahmet and Sun, Chen and Wang, Zirui and Chang, Kai-Wei and Sun, Yizhou and Schmid, Cordelia and Ross, David A and Fathi, Alireza},
  booktitle=CVPR,
  xxxxpages={23369--23379},
  year={2023}
}

@inproceedings{realm,
  title={Retrieval augmented language model pre-training},
  author={Guu, Kelvin and Lee, Kenton and Tung, Zora and Pasupat, Panupong and Chang, Mingwei},
  booktitle=ICML,
  xxxxpages={3929--3938},
  year={2020}
}

@article{lazaridou2022internet,
  title={Internet-augmented language models through few-shot prompting for open-domain question answering},
  author={Lazaridou, Angeliki and Gribovskaya, Elena and Stokowiec, Wojciech and Grigorev, Nikolai},
  journal={arXiv preprint arXiv:2203.05115},
  year={2022}
}

@inproceedings{clip,
  title={Learning transferable visual models from natural language supervision},
  author={Radford, Alec and Kim, Jong Wook and Hallacy, Chris and Ramesh, Aditya and Goh, Gabriel and Agarwal, Sandhini and Sastry, Girish and Askell, Amanda and Mishkin, Pamela and Clark, Jack and others},
  booktitle=ICML,
  xxxxpages={8748--8763},
  year={2021}
}

@software{yolov8,
  author = {Glenn Jocher and Ayush Chaurasia and Jing Qiu},
  title = {Ultralytics YOLOv8},
  version = {8.0.0},
  year = {2023},
  orcid = {0000-0001-5950-6979, 0000-0002-7603-6750, 0000-0003-3783-7069},
  license = {AGPL-3.0}
}

@inproceedings{sam,
  title={Segment anything},
  author={Kirillov, Alexander and Mintun, Eric and Ravi, Nikhila and Mao, Hanzi and Rolland, Chloe and Gustafson, Laura and Xiao, Tete and Whitehead, Spencer and Berg, Alexander C and Lo, Wan-Yen and others},
  booktitle=ICCV,
  xxxxpages={4015--4026},
  year={2023}
}

@inproceedings{refcoco,
  title={Modeling context in referring expressions},
  author={Yu, Licheng and Poirson, Patrick and Yang, Shan and Berg, Alexander C and Berg, Tamara L},
  booktitle=ECCV,
  xxxxpages={69--85},
  year={2016}
}

@inproceedings{referit,
  author       = {Sahar Kazemzadeh and
                  Vicente Ordonez and
                  Mark Matten and
                  Tamara L. Berg},
  editor       = {Alessandro Moschitti and
                  Bo Pang and
                  Walter Daelemans},
  title        = {ReferItGame: Referring to Objects in Photographs of Natural Scenes},
  booktitle    = EMNLP,
  xxxxpages        = {787--798},
  year         = {2014},
}

@inproceedings{sesame,
  title={See Say and Segment: Teaching LMMs to Overcome False Premises},
  author={Wu, Tsung-Han and Biamby, Giscard and Chan, David and Dunlap, Lisa and Gupta, Ritwik and Wang, Xudong and Gonzalez, Joseph E and Darrell, Trevor},
  booktitle=CVPR,
  xxxxpages={13459--13469},
  year={2024}
}

@inproceedings{ragsurvey,
  title={A survey on rag meeting llms: Towards retrieval-augmented large language models},
  author={Fan, Wenqi and Ding, Yujuan and Ning, Liangbo and Wang, Shijie and Li, Hengyun and Yin, Dawei and Chua, Tat-Seng and Li, Qing},
  booktitle={Proceedings of the 30th ACM SIGKDD Conference on Knowledge Discovery and Data Mining},
  xxxxpages={6491--6501},
  year={2024}
}

@inproceedings{yu2022generate,
  title={Generate rather than retrieve: Large language models are strong context generators},
  author={Yu, Wenhao and Iter, Dan and Wang, Shuohang and Xu, Yichong and Ju, Mingxuan and Sanyal, Soumya and Zhu, Chenguang and Zeng, Michael and Jiang, Meng},
  booktitle=ICLR,
  year={2022}
}

@article{ding2025multimodal,
  title={Multimodal referring segmentation: A survey},
  author={Ding, Henghui and Tang, Song and He, Shuting and Liu, Chang and Wu, Zuxuan and Jiang, Yu-Gang},
  journal={arXiv preprint arXiv:2508.00265},
  year={2025}
}
}

\end{document}